\documentclass[10pt,journal,compsoc]{IEEEtran}
\ifCLASSOPTIONcompsoc
  \usepackage[nocompress]{cite}
\else
  \usepackage{cite}
\fi
\ifCLASSINFOpdf
\else
\fi
\usepackage{booktabs}
\usepackage{amsmath,amssymb,amsfonts}
\usepackage{graphicx}
\usepackage[hyphens]{url}
\usepackage{comment}
\begin{document}
\title{Explainable Artificial Intelligence Approaches: A Survey}
\author{Sheikh Rabiul Islam,~University of Hartford,
       William Eberle,~Tennessee Tech University,
       Sheikh Khaled Ghafoor,~Tennessee Tech University, Mohiuddin Ahmed,~Edith Cowan University% <-this % stops a %%space
%%\IEEEcompsocitemizethanks{\IEEEcompsocthanksitem M. Shell was with the Department
%%of Electrical and Computer Engineering, Georgia Institute of Technology, Atlanta,
%%GA, 30332.\protect\\
% note need leading \protect in front of \\ to get a newline within \thanks as
% \\ is fragile and will error, could use \hfil\break instead.
%%E-mail: see http://www.michaelshell.org/contact.html
%%IEEEcompsocthanksitem J. Doe and J. Doe are with Anonymous University.}% <-this % stops an unwanted %%space
%%\thanks{Manuscript received April 19, 2005; revised August 26, 2015.}
}

% The paper headers
%%\markboth{Journal of \LaTeX\ Class Files,~Vol.~14, No.~8, August~2015}%
%%{Shell \MakeLowercase{\textit{et al.}}: Bare Demo of IEEEtran.cls for Computer Society Journals}

\IEEEtitleabstractindextext{%
\begin{abstract}
The lack of explainability of a decision from an Artificial Intelligence (AI) based ``black box'' system/model, despite its superiority in many real-world applications, is a key stumbling block for adopting AI in many high stakes applications of different domain or industry. While many popular Explainable Artificial Intelligence (XAI) methods or approaches are available to facilitate a human-friendly explanation of the decision, each has its own merits and demerits, with a plethora of open challenges. We demonstrate popular XAI methods with a mutual case study/task (i.e., credit default prediction), analyze for competitive advantages from multiple perspectives (e.g., local, global),  provide meaningful insight on quantifying explainability, and recommend paths towards responsible or human-centered AI using XAI as a medium. Practitioners can use this work as a catalog to understand, compare, and correlate competitive advantages of popular XAI methods. In addition, this survey elicits  future research directions towards responsible or human-centric AI systems, which is crucial to adopt AI in high stakes applications.
\end{abstract}

% Note that keywords are not normally used for peerreview papers.
}

% make the title area
\maketitle
\IEEEdisplaynontitleabstractindextext
\IEEEpeerreviewmaketitle

\IEEEraisesectionheading{\section{Introduction}\label{sec:introduction}}
\IEEEPARstart{A}{rtificial} Intelligence (AI) has become an integral part of many real-world applications. Factors fueling the proliferation of AI-based algorithmic decision making in many disciplines include: (1) the demand for processing a variety of voluminous data, (2) the availability of powerful computing resources (e.g., GPU computing, cloud computing), and (3) powerful and new algorithms. However, most of the successful AI-based models are ``black box'' in nature, making it a challenge to understand how the model or algorithm works and generates decisions. In addition, the decisions from AI systems affect human interests, rights, and lives; consequently, the decision is crucial for high stakes applications such as credit approval in finance, automated machines in defense, intrusion detection in cybersecurity, etc. Regulators are introducing new laws such as European Union's General Data Protection Regulation (GDPR) \footnote{https://www.eugdpr.org} \cite{ras2018explanation} aka  ``right to explanation'' \cite{goodman2016eu}, US government's ``Algorithmic Accountability Act of 2019'' \footnote{https://www.senate.gov} \cite{algorithmic_accountability}, or U.S. Department of Defense's Ethical Principles for Artificial Intelligence  \footnote{https://www.defense.gov} \cite{dod_ai_ethical_principle} ) to tackle primarily fairness, accountability, and transparency-related risks with automated decision making systems. 

XAI is a re-emerging research trend, as the need to advocate these principles/laws, and promote the explainable decision-making system and research, continues to increase. Explanation systems were first introduced in the early '80s to explain the decisions of expert systems. Later, the focus of the explanation systems shifted towards human-computer systems (e.g., intelligent tutoring systems) to provide better cognitive support to users. The primary reason for the renewed interest in XAI research has stemmed from recent advancements in AI and ML, and their application to a wide range of areas, as well as prevailing concerns over the unethical use, lack of transparency, and undesired biases in the models. Many real-world applications in the Industrial Control System (ICS) greatly increase the efficiency of industrial production from the automated equipment and production processes \cite{wang2020explaining}. However, in this setting, the use of 'black box' is still not in a favorable position due to the lack of explainability and transparency of the model and decisions.  

According to \cite{samek2017explainable} and \cite{fernandez2019evolutionary}, XAI encompasses Machine Learning (ML) or AI systems/tools for demystifying black models internals (e.g., what the models have learned) and/or for explaining individual predictions. In general, explainability of an AI model's prediction is the extent of transferable qualitative understanding of the relationship between model input and prediction (i.e., selective/suitable causes of the event) in a recipient friendly manner. The term ``explainability'' and ``interpretability'' are being used interchangeably throughout the literature. To this end, in the case of an intelligent system (i.e., AI-based system), it is evident that explainability is more than interpretability in terms of importance, completeness, and fidelity of prediction. Based on that, we will use these terms accordingly where appropriate. 

Due to the increasing number of XAI approaches, it has become challenging to understand the pros, cons, and competitive advantages, associated with the different domains. In addition, there are lots of variations among different XAI methods, such as whether a method is global (i.e., explains the model's behavior on the entire data set), local (i.e., explains the prediction or decision of a particular instance), ante-hoc (i.e. involved in the pre training stage), post-hoc (i.e. works on already trained model), or surrogate (i.e. deploys a simple model to emulate the prediction of a ``black box'' model). However, despite many reviews on XAI methods, there is still a lack of comprehensive analysis of XAI when it comes to these methods and perspectives.  

Some of the popular work/tools on XAI are LIME, DeepVis Toolbox, TreeInterpreter, Keras-vis, Microsoft InterpretML, MindsDB, SHAP, Tensorboard WhatIf, Tensorflow's Lucid, Tensorflow's Cleverhans, etc. However, a few of these work/tools are model specific. For instance, DeepVis, keras-vis, and Lucid are for a neural network's explainability, and TreeInterpreter is for a tree-based model's explainability. At a high level, each of the proposed approaches have similar concepts, such as feature importance, feature interactions, shapely values, partial dependence, surrogate models, counterfactual, adversarial, prototypes and knowledge infusion. However, despite some visible progress in XAI methods, the quantification or evaluation of explainability is under-focused, and in particular, when it comes to human study-based evaluations.  

In this paper, we (1) demonstrate popular methods/approaches towards XAI with a mutual task (i.e., credit default prediction) and explain the working mechanism in layman's terms, (2) compare the pros, cons, and competitive advantages of each approach with their associated challenges, and analyze those from multiple perspectives (e.g., global vs local, post-hoc vs ante-hoc, and inherent vs emulated/approximated explainability), (3) provide meaningful insight on quantifying explainability, and (4) recommend a path towards responsible or human-centered AI using XAI as a medium.  Our survey is only one among the recent ones (See Table \ref{tab:survey}) which includes a mutual test case with useful insights on popular XAI methods (See Table \ref{tab:interpretability_methods}).

\begin{table}[!htbp]
\caption{Comparison with other Surveys} % title of Table
\label{tab:survey}
\centering % used for centering table
\begin{tabular}{ccc} % centered columns (7 columns)
\hline  %inserts double horizontal lines
\bf{Survey} & \bf{Reference} &  \bf{Mutual test case}   \\ [1.0ex] % inserts table
\hline   % inserts single horizontal line
Adadi et al., 2018 & \cite{adadi2018peeking} & $\times$  \\
Mueller et al., 2019 & \cite{mueller2019explanation} & $\times$  \\
Samek et al., 2017 & \cite{samek2017explainable} & $\times$  \\
Molnar et al., 2019 & \cite{molnar2019quantifying} & $\times$  \\
Staniak et al., 2018 & \cite{staniak2018explanations} & $\times$  \\
Gilpin et al., 2018 & \cite{gilpin2018explaining} & $\times$  \\
Collaris et al., 2018 & \cite{collaris2018instance} & $\times$  \\
Ras et al., 2018 & \cite{ras2018explanation} & $\times$  \\
Dosilovic et al., 2018 & \cite{dovsilovic2018explainable} & $\times$  \\
Tjoa et al., 2019 & \cite{tjoa2019survey} & $\times$  \\
Dosi-Valez et al., 2017 & \cite{doshi2017towards } & $\times$  \\
Rudin et al., 2019 & \cite{rudin2019stop} & $\times$  \\
Arrieta et al., 2020 & \cite{arrieta2020explainable} & $\times$  \\
Miller et al., 2018 & \cite{miller2018explanation} & $\times$  \\
Zhang et al., 2018 & \cite{zhang2018visual} & $\times$  \\
{\bf{This Survey}} &  & $\checkmark$ \\
\hline %inserts single l
\end{tabular}

\end{table}

We start with a background of related works (Section \ref{sec:background}), followed by a description of the test case in Section \ref{sec:testcase}, and then a review of XAI methods in Section \ref{sec:xai_methods}. We conclude with an overview of quantifying explainability and a discussion addressing open questions and future research directions towards responsible for human-centered AI in Section \ref{sec:challenge_future_research}.

%
%
%
%###################################  BACKGROUND ############################ 
%
%
%
\section{Background} \label{sec:background}
Research interests in XAI are re-emerging. The earlier works such as \cite{chandrasekaran1989explaining}, \cite{swartout1993explanation}, and \cite{swartout1985rule} focused primarily on explaining the decision process of knowledge-based systems and expert systems. The primary reason behind the renewed interest in XAI research has stemmed from the recent advancements in AI, its application to a wide range of areas, the concerns over unethical use, lack of transparency, and undesired biases in the models. In addition, recent laws by different governments are necessitating more research in XAI. According to \cite{samek2017explainable} and \cite{fernandez2019evolutionary}, XAI encompasses Machine Learning (ML) or AI systems for demystifying black models internals (e.g., what the models have learned) and/or for explaining individual predictions.     

In 2019, Mueller et al. presents a comprehensive review of the approaches taken by a number of types of ``explanation systems'' and characterizes those into three generations: (1) first-generation systems\textemdash{for instance, expert systems from the early 70's}, (2) second generation systems\textemdash{for instance, intelligent tutoring systems}, and (3) third generation systems\textemdash{tools and techniques from the recent renaissance starting from 2015}\cite{mueller2019explanation}. The first generation systems attempt to clearly express the internal working process of the system by embedding expert knowledge in rules often elicited directly from experts (e.g., via transforming rules into natural language expressions). The second generations systems can be regarded as the human-computer system designed around human knowledge and reasoning capacities to provide cognitive support. For instance, arranging the interface in such a way that complements the knowledge that the user is lacking. Similar to the first generation systems, the third generation systems also attempt to clarify the inner workings of the systems. But this time, these systems are mostly ``black box'' (e.g., deep nets, ensemble approaches). In addition, nowadays, researchers are using advanced computer technologies in data visualizations, animation, and video, that have a strong potential to drive the XAI research further. Many new ideas have been proposed for generating explainable decisions from the need of primarily accountable, fair, and trust-able systems and decisions.  

There has been some previous work  \cite{molnar2019quantifying} that mentions three notions for quantification of explainability. Two out of three notions involve experimental studies with humans (e.g., domain expert or a layperson, that mainly investigate whether a human can predict the outcome of the model) \cite{dhurandhar2017tip}, \cite{friedler2019assessing}, \cite{huysmans2011empirical}, \cite{poursabzi2018manipulating}, \cite{zhou2018measuring}. The third notion (proxy tasks) does not involve a human, and instead uses known truths as a metric (e.g., the less the depth of the decision tree, the more explainable the model).

Some mentionable reviews on XAI are listed in Table \ref{tab:survey}. However, while these works provide analysis from one or more of the mentioned perspectives, a comprehensive review considering all of the mentioned important perspectives, using a mutual test case, is still missing. Therefore, we attempt to provide an overview using a demonstration of a mutual test case or task, and then analyze the various approaches from multiple perspectives, with some future directions of research towards responsible or human-centered AI.  

%
%
%
%################################### Test Case ############################ 
%
%
%

\section{Test Case} \label{sec:testcase}
The mutual test case or task that we use in this paper to demonstrate and evaluate the XAI methods is \textit{credit default prediction}. This mutual test case enables a better understanding of the comparative advantages of different XAI approaches. We predict whether a customer is going to default on a mortgage payment (i.e., unable to pay monthly payment) in the near future or not, and explain the decision using different XAI methods in a human-friendly way. We use the popular Freddie Mac \cite{freddie_mac} dataset for the experiments. Table \ref{tab:dataset} lists some important features and their descriptions. The description of features are taken from the data set's \cite{freddie_mac} user guide.

\begin{table*}[h!]
\centering
\caption{Dataset description}
\label{tab:dataset}
\begin{tabular}{ll}
\toprule
{\bf{Feature}}                      & {\bf{Description}}                                                                                          \\
\midrule
creditScore                  & A number in between 300 and 850 that indicates the  creditworthiness of the borrowers.                       \\
originalUPB                  & Unpaid principle balance on the note date.                                                           \\
originalInterestRate         & Original interest rate as indicated by the mortgage note.                                            \\
currentLoanDelinquencyStatus & Indicates the number of days the borrower is delinquent.                                             \\
numberOfBorrower             & Number of borrower who are obligated to repay the loan.                                              \\
currentInterestRate          & Active interest rate on the note.                                                                         \\
originalCombinedLoanToValue  & Ratio of all mortgage loans  and apprised price of mortgaged property on the note date.  \\
currentActualUPB             & Unpaid principle balance as of latest month of payment.                                              \\
defaulted                    & Whether the customer was default on payment (1) or not (0.) \\
\bottomrule
\end{tabular}
\end{table*}
We use well-known programming language R's package ``iml'' \cite{iml} for producing the results for the XAI methods described in this review.  

%
%
%
%################################### Explainable Artificial Intelligence Methods ############################ 
%
%
%
\section{Explainable Artificial Intelligence Methods} \label{sec:xai_methods}
This section summarizes different explainability methods with their pros, cons, challenges, and competitive advantages primarily based on two recent comprehensive surveys: \cite{molnar2018interpretable} and \cite{doshi2017towards}. We then enhance the previous surveys with a multi-perspective analysis, recent research progresses, and future research directions.\cite{doshi2017towards} broadly categorize methods for explanations into three kinds: Intrinsically Interpretable Methods, Model Agnostic Methods, and Example-Based Explanations.

\subsection{Intrinsically Interpretable Methods}
The convenient way to achieve explainable results is to stick with intrinsically interpretable models such as Linear Regression, Logistic Regression, and Decision Trees by avoiding the use of ``black  box’’ models. However, usually, this natural explainability comes with a cost in performance.  

In a \textbf{Linear Regression}, the predicted target consists of the weighted sum of input features. So the weight or coefficient of the linear equation can be used as a medium of explaining prediction when the number of features is small. 
\begin{equation}
\label{eq:lr}
    y = b\textsubscript{0} + b\textsubscript{1}*x\textsubscript{1} + ... + b\textsubscript{n}*x\textsubscript{n} + \epsilon
\end{equation}
In Formula \ref{eq:lr}, y is the target (e.g., chances of credit default), b\textsubscript{0} is a constant value known as the intercept (e.g., .33), b\textsubscript{i} is the learned feature's weight or coefficient (e.g., .33) for the corresponding feature x\textsubscript{i} (e.g., credit score), and $\epsilon$ is a constant error term (e.g., .0001).  Linear regression comes with an interpretable linear relationship among features. However, in cases where there are multiple correlated features, the distinct feature influence becomes indeterminable as the individual influences in prediction are not additive to the overall prediction anymore.

\textbf{Logistic Regression} is an extension of Linear Regression to the classification problems. It models the probabilities for classification tasks. The interpretation of Logistic Regression is different from Linear Regression as it gives a probability between 0 and 1, where the weight might not exactly represent the linear relationship with the predicted probability. However, the weight provides an indication of the direction of influence (negative or positive) and a factor of influence between classes, although it is not additive to the overall prediction. 

\begin{figure}[h!]
    \centering
    \includegraphics[width=\linewidth]{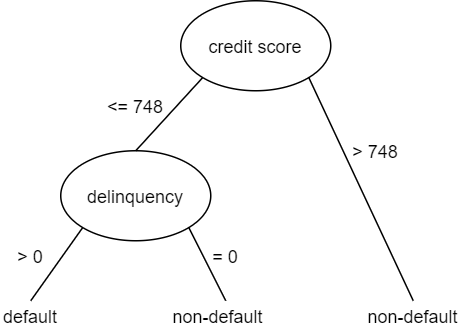}
    \caption{Decision Trees}
    \label{fig:dt}
\end{figure}

\textbf{Decision Tree-based models} split the data multiple times based on a cutoff threshold at each node until it reaches a leaf node. Unlike Logistic and Linear Regression, it works even when the relationship between input and output is non-linear, and even when the features interact with one another (i.e., a correlation among features). In a Decision Tree, a path from the root node (i.e., starting node) (e.g., credit score in Figure \ref{fig:dt}) to a leaf node (e.g., default) tells how the decision (the leaf node) took place. Usually, the nodes in the upper-level of the tree have higher importance than lower-level nodes. Also, the less the number of levels (i.e., height) a tree has, the higher the level of explainability the tree possesses. In addition, the cutoff point of a node in the Decision Trees provides counterfactual information\textemdash{for instance, increasing the value of a feature equal to the cutoff point will reverse the decision/prediction}. In Figure \ref{fig:dt}, if the credit score is greater than the cutoff point 748, then the customer is predicted as non-default. Also, tree-based explanations are contrastive, i.e., a ''what if'' analysis provides the relevant alternative path to reach a leaf node. According to the tree in Figure \ref{fig:dt}, there are two separate paths (credit score $\rightarrow$ delinquency $\rightarrow$ non-default; and credit score $\rightarrow$ non-default) that lead to a non-default classification.  

However, tree-based explanations cannot express the linear relationship between input features and output. It also lacks smoothness; slight changes in input can have a big impact on the predicted output. Also, there can be multiple different trees for the same problem. Usually, the more the nodes or depth of the tree, the more challenging it is to interpret the tree.

\textbf{Decision Rules} (simple IF-THEN-ELSE conditions) are also an inherent explanation model. For instance, ''IF credit score is less than or equal to 748 AND if the customer is delinquent on payment for more than zero days (condition), THEN the customer will default on payment (prediction)''. Although IF-THEN rules are straightforward to interpret, it is mostly limited to classification problems (i.e., does not support a regression problem), and inadequate in describing linear relationships. In addition, the \textbf{RuleFit} algorithm \cite{friedman2008predictive} has an inherent interpretation to some extent as it learns sparse linear models that can detect the interaction effects in the form of decision rules.  Decision rules consist of the combination of split decisions from each of the decision paths. However, besides the original features, it also learns some new features to capture the interaction effects of original features. Usually, interpretability degrades with an increasing number of features. 

Other interpretable models include the extension of linear models such as \textbf{Generalized Linear Models (GLMs)} and \textbf{Generalized Additive Models (GAMs)}; they help to deal with some of the assumptions of linear models (e.g., the target outcome y and given features follow a Gaussian Distribution; and no interaction among features). However, these extensions make models more complex (i.e., added interactions) as well as less interpretable. In addition, a \textbf{Naïve Bayes Classifier} based on Bayes Theorem, where the probability of classes for each of the features is calculated independently (assuming strong feature independence), and \textbf{K-Nearest Neighbors}, which uses nearest neighbors of a data point for prediction (regression or classification), also fall under intrinsically interpretable models.

\subsection{Model-Agnostic Methods} 
Model-agnostic methods separate explanation from a machine learning model, allowing the explanation method to be compatible with a variety of models. This separation has some clear advantages such as (1) the interpretation method can work with multiple ML models, (2) provides different forms of explainability (e.g., visualization of feature importance, linear formula) for a particular model, and (3) allows for a flexible representation\textemdash{a text classifier uses abstract word embedding for classification but uses actual words for explanation}. Some of the model-agnostic interpretation methods include Partial Dependence Plot (PDP), Individual Conditional Expectation (ICE), Accumulation Local Effects (ALE) Plot, Feature Interaction, Feature Importance, Global Surrogate, Local Surrogate (LIME), and Shapley Values (SHAP).
 
\subsubsection{Partial Dependence Plot (PDP)}
\begin{figure}[h!]
    \centering
    \includegraphics[width=\linewidth]{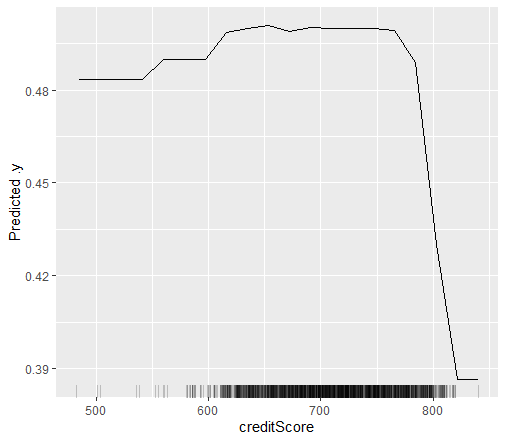}
    \caption{Partial Dependence Plot (PDP)}
    \label{fig:pdp}
\end{figure}

The partial Dependence Plot (PDP) or PD plot shows the marginal effect of one or two features (at best three features in 3-D) on the predicted outcome of an ML model \cite{friedman2001greedy}. It is a global method, as it shows an overall model behavior, and is capable of showing the linear or complex relationships between target and feature(s). It provides a function that depends only on the feature(s) being plotted by marginalizing over other features in such a way that includes the interactions among them. PDP provides a clear and causal interpretation by providing the changes in prediction due to changes in particular features. However, PDP assumes features under the plot are not correlated with the remaining features. In the real world, this is unusual. Furthermore, there is a practical limit of only two features that PD plot can clearly explain at a time. Also, it is a global method, as it plots the average effect (from all instances) of a feature(s) on the prediction, and not for all features on a specific instance. The PD plot in Figure \ref{fig:pdp} shows the effect of credit score on prediction. Individual bar lines along the X axis represent the frequency of samples for different ranges of credit scores.

\subsubsection{Individual Conditional Expectation (ICE)}
\begin{figure}[h!]
    \centering
    \includegraphics[width=\linewidth]{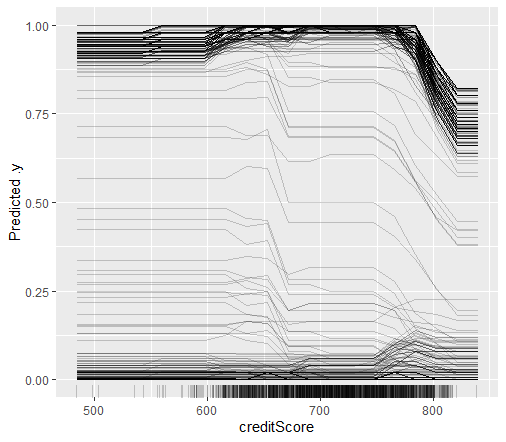}
    \caption{Individual Conditional Expectation (ICE)}
    \label{fig:ice}
\end{figure}

\begin{figure}[h!]
    \centering
    \includegraphics[width=\linewidth]{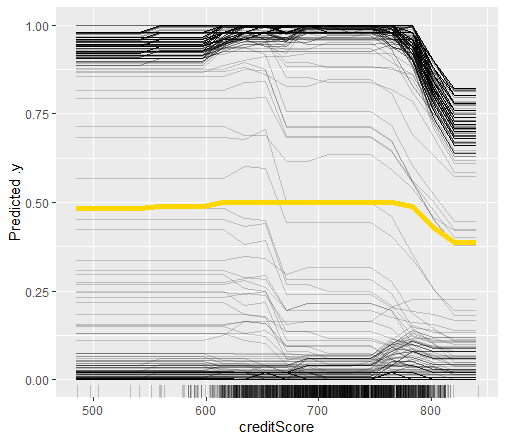}
    \caption{PDP and ICE combined together in the same plot}
    \label{fig:pdp_ice}
\end{figure}

Unlike PDP, ICE plots one line per instance showing how a feature influences the changes in prediction (See Figure \ref{fig:ice}. The average on all lines of an ICE plot gives a PD plot \cite{goldstein2017package} (i.e., the single line shown in the PD plot in Figure \ref{fig:pdp}). Figure \ref{fig:pdp_ice}, combines both PDP and ICE together for a better interpretation.    

Although ICE curves are more intuitive to understand than a PD plot, it can only display one feature meaningfully at a time. In addition, it also suffers from the problem of correlated features and overcrowded lines when there are many instances.

\subsubsection{Accumulated Local Effects (ALE) Plot}
\begin{figure}[h!]
    \centering
    \includegraphics[width=\linewidth]{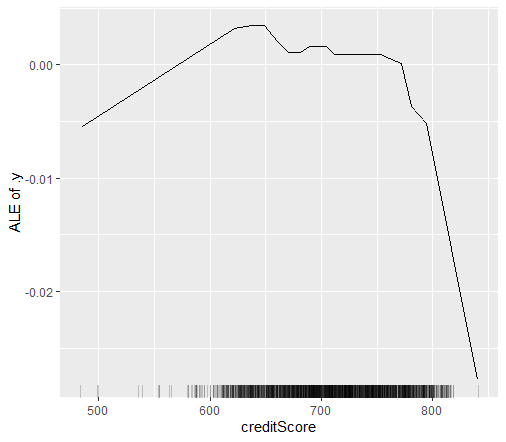}
    \caption{Accumulated Local Effects (ALE) Plot}
    \label{fig:ale}
\end{figure}

Similar to PD plots (Figure \ref{fig:pdp}, ALE plots (Figure \ref{fig:ale} describe how features influence the prediction on average. However, unlike PDP, ALE plot reasonably works well with correlated features and is comparatively faster. Although ALE plot is not biased to the correlated features, it is challenging to interpret the changes in prediction when features are strongly correlated and analyzed in isolation. In that case, only plots showing changes in both correlated features together make sense to understand the changes in the prediction.

\subsubsection{Feature Interaction}
\begin{figure}[h!]
    \centering
    \includegraphics[width=\linewidth]{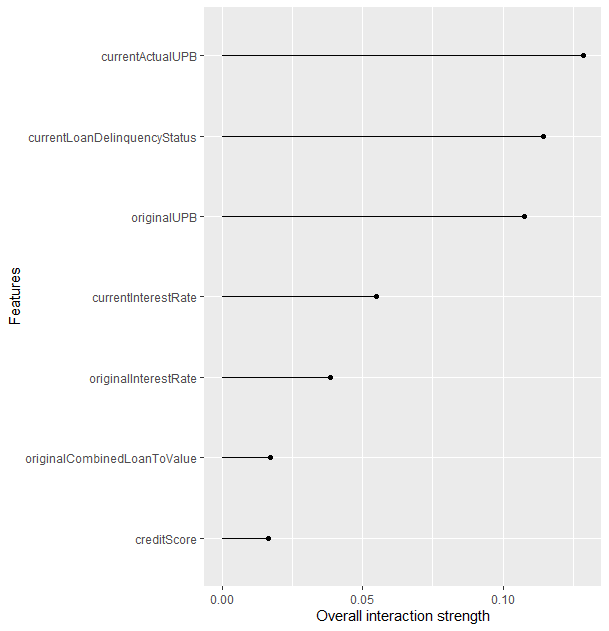}
    \caption{Feature interaction}
    \label{fig:feature_interaction}
\end{figure}

When the features interact with one another, individual feature effects do not sum up to the total feature effects from all features combined. An H-statistic (i.e., Friedman's H-statistic) helps to detect different types of interaction, even with three or more features. The interaction strength between two features is the difference between the \textit{partial dependence function for those two features together} and the sum of the \textit{partial dependence functions for each feature separately}. Figure \ref{fig:feature_interaction} shows the interaction strength of each participating feature. For example,  \textit{current Actual UPB} has the highest level of interaction with other features, and \textit{credit score} has the least interaction with other features. However, calculating feature interaction is computationally expensive. Furthermore, using sampling instead of the entire dataset usually shows variances from run to run.  \ref{fig:feature_interaction},  

\subsubsection{Feature Importance}
\begin{figure}[h!]
    \centering
    \includegraphics[width=\linewidth]{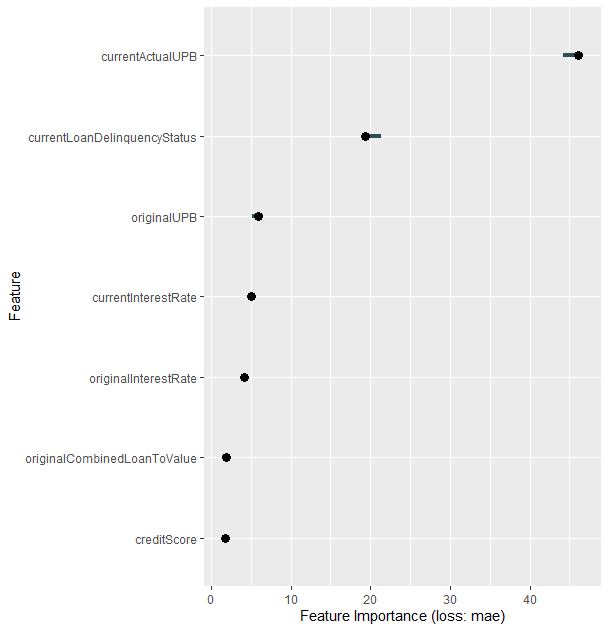}
    \caption{Feature importance}
    \label{fig:feature_importance}
\end{figure}

Usually, the feature importance of a feature is the increase in the prediction error of the model when we permute the values of the feature to break the true relationship between the feature and the true outcome.  After shuffling the values of the feature, if errors increase, then the feature is important.  \cite{breiman2001random} introduced the permutation-based feature importance for Random Forests; later \cite{fisher2018model} extended the work to a model-agnostic version. Feature importance provides a compressed and global insight into the ML model’s behavior. For example, Figure \ref{fig:feature_importance} shows the importance of each participating feature, \textit{current Actual UPB} possess the highest feature importance, and \textit{credit score} possess the lowest feature importance. Although feature importance takes into account both the main feature effect and interaction, this is a disadvantage as feature interaction is included in the importance of correlated features. We can see that the feature \textit{current Actual UPB} possesses the highest feature importance (Figure \ref{fig:feature_importance}), at the same time it also possesses the highest interaction strength \ref{fig:feature_interaction}. As a result, in the presence of interaction among features, the feature importance does not add up to total drop-in of performance. Besides, it is unclear whether the test set or training set should be used for feature importance, as it demonstrates variance from run to run in the shuffled dataset. It is necessary to mention that feature importance also falls under the global methods.

\subsubsection{Global Surrogate}
\begin{figure}[h!]
    \centering
    \includegraphics[width=\linewidth]{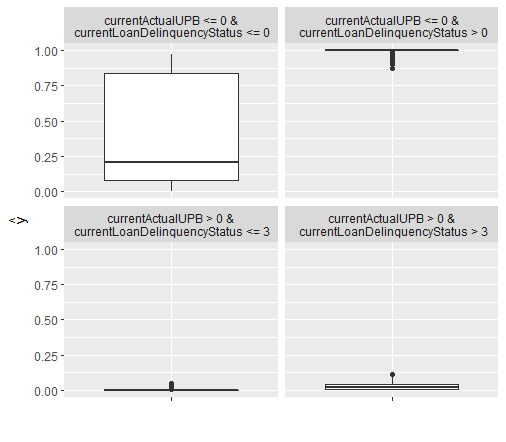}
    \caption{Global surrogate}
    \label{fig:global_surrogate}
\end{figure}

A global surrogate model tries to approximate the overall behavior of a ``black box’’ model using an interpretable ML model. In other words, surrogate models try to approximate the prediction function of a black-box model using an interpretable model as correctly as possible, given the prediction is interpretable. It is also known as a meta-model, approximate model, response surface model, or emulator. We approximate the behavior of a Random Forest using CART decision trees (Figure  \ref{fig:global_surrogate}). The original black box model could be avoided given the surrogate model demonstrates a comparable performance. Although a surrogate model comes with interpretation and flexibility (i.e., such as model agnosticism), diverse explanations for the same ``black box’’ such as multiple possible decision trees with different structures, is a drawback. Besides, some would argue that this is only an illusion of interpretability.

\subsubsection{Local Surrogate (LIME)}
\begin{figure}[h!]
    \centering
    \includegraphics[width=\linewidth]{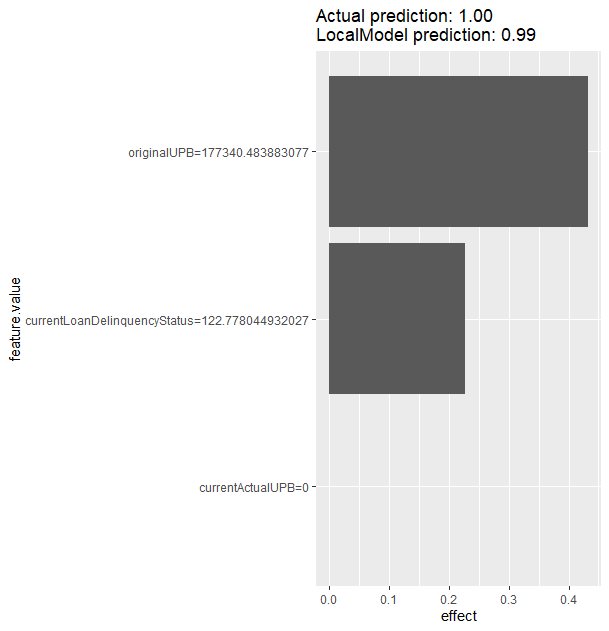}
    \caption{Local Interpretable Model-Agnostic Explanations (LIME)}
    \label{fig:lime}
\end{figure}

Unlike global surrogate, local surrogate explains individual predictions of black-box models. Local Interpretable Model-Agnostic Explanations (LIME) was proposed by \cite{ribeiro2016should}. Lime trains an inherently interpretable model (e.g., Decision Trees) on a new dataset made from the permutation of samples and the corresponding prediction of the black box. Although the learned model can have a good approximation of local behavior, it does not have a good global approximation. This trait is also known as local fidelity. Figure \ref{fig:lime} is a visualization of the output from LIME. For a random sample, the black box predicts that a customer will default on payment with a probability of 1; the local surrogate model, LIME also predict that the customer will default on the payment, however, the probability is 0.99, that is little less than the black box models prediction. LIME also shows which feature contributes to the decision making and by how much. Furthermore, LIME allows replacing the underlying ``black box’’ model by keeping the same local interpretable model for the explanation. In addition, LIME works for tabular data, text, and images. As LIME is an approximation model, and the local model might not cover the complete attribution due to the generalization (e.g., using shorter trees, lasso optimization), it might be unfit for cases where we legally need complete explanations of a decision. Furthermore, there is no consensus on the boundary of the neighborhood for the local model; sometimes, it provides very different explanations for two nearby data points.

\subsubsection{Shapley Values}
\begin{figure*}[h!]
    \centering
    \includegraphics[width=.68\linewidth]{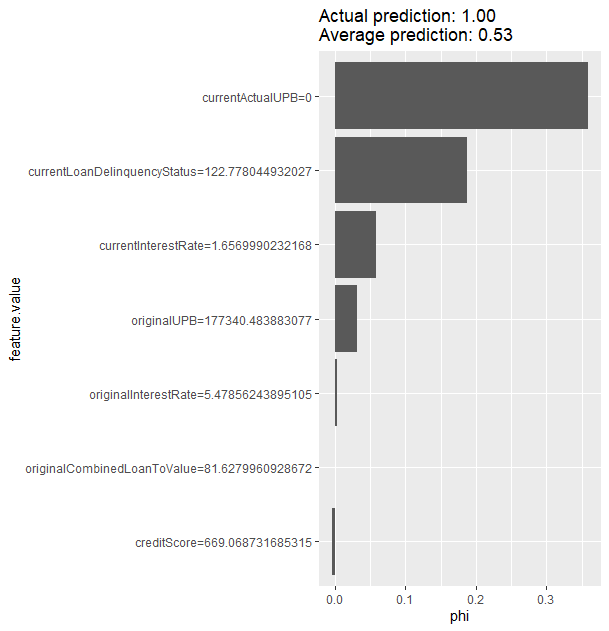}
    \caption{Shapely values}
    \label{fig:shapely}
\end{figure*}

Shapley is another local explanation method. In 1953, Shapley \cite{shapley1953value} coined the Shapley Value. It is based on coalitional game theory that helps to distribute feature importance among participating features fairly. Here the assumption is that each feature value of the instance is a player in a game, and the prediction is the overall payout that is distributed among players (i.e., features) according to their contribution to the total payout (i.e., prediction). We use Shapely values (See Figure \ref{fig:shapely}) to analyze the prediction of a random forest model for the credit default prediction problem. The actual prediction for a random sample is 1.00, the average prediction from all samples in the data set is 0.53, and their difference .47 (1.00 $-$ 0.53) consists of the individual contributions from the features (e.g., \textit{Current Actual UPB} contributes 0.36). The Shapely Value is the average contribution in prediction over all possible coalition of features, which make it computationally expensive when there is a large number of features\textemdash{for example, for k number of features, there will be 2\textsuperscript{k} number of coalitions}.  Unlike LIME, Shapely Value is an explanation method with a solid theory that provides full explanations. However, it also suffers from the problem of correlated features. Furthermore, the Shapely value returns a single value per feature; there is no way to make a statement about the changes in output resulting from the changes in input. One mentionable implementation of the Shapely value is in the work of \cite{lundberg2016unexpected} that they call SHAP.

\subsubsection{Break Down}
\begin{figure*}[h!]
    \centering
    \includegraphics[width=\linewidth]{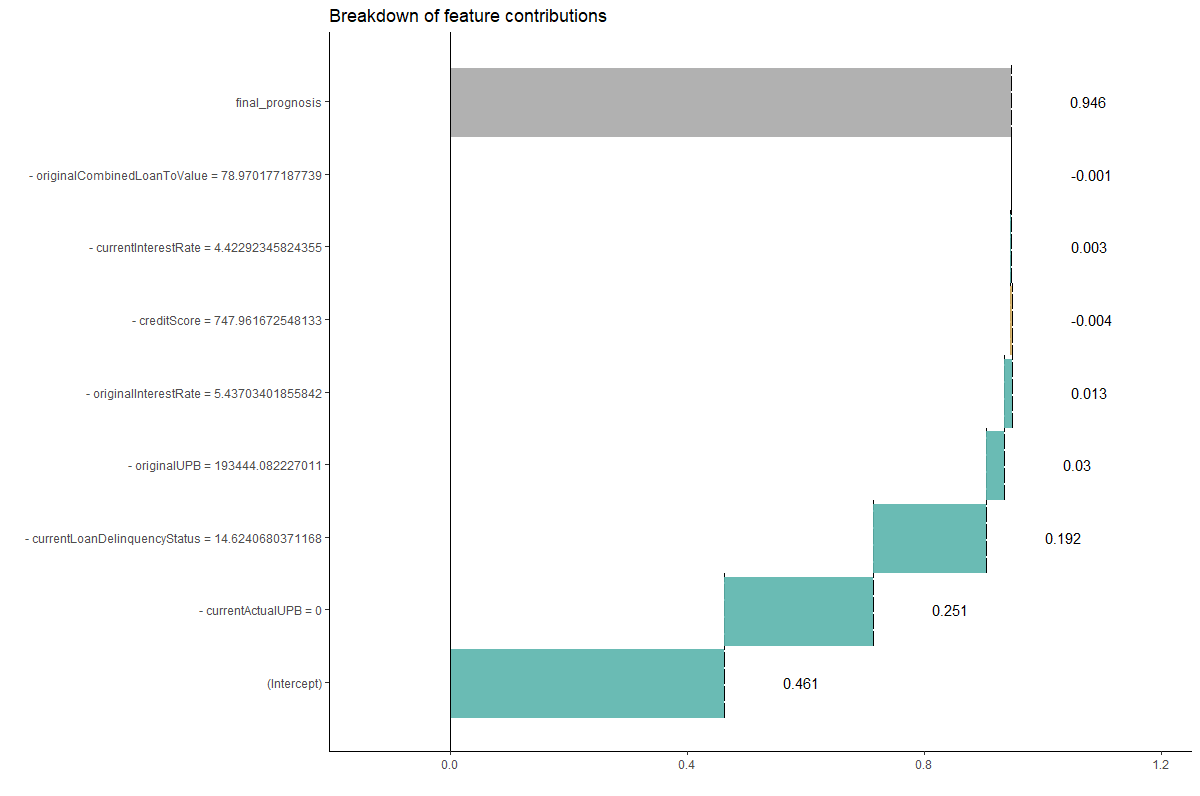}
    \caption{Breakdown}
    \label{fig:breakdown}
\end{figure*}

The Break Down package provides the local explanation and is loosely related to the partial dependence algorithm with an added step-wise procedure known as ``Break Down'' (proposed by \cite{staniak2018explanations}). It uses a greedy strategy to identify and remove features iteratively based on their influence on the overall average predicted response (baseline) \cite{Chapter179:online}. For instance, from the game theory perspective, it starts with an empty team, then adds feature values one by one based on their decreasing contribution. In each iteration, the amount of contribution from each feature depends on the features values of those are already in the team, which is considered as a drawback of this approach. However, it is faster than the Shapley value method due to the greedy approach, and for models without interactions, the results are the same \cite{molnar2018interpretable}. Figure \ref{fig:breakdown} is a visualization of \textit{break down} for a random sample, showing contribution (positive or negative) from each of the participating features towards the final prediction.   

\subsection{Example-Based Explanations}
Example-Based Explanation methods use particular instances from the dataset to explain the behavior of the model and the distribution of the data in a model agnostic way. It can be expressed as ``X is similar to Y and Y caused Z, so the prediction says X will cause Z''. According to \cite{molnar2018interpretable}, a few explanation methods that fall under Example-Based Explanations are described as follows:

\subsubsection{Counterfactual}
The counterfactual method indicates the required changes in the input side that will have significant changes (e.g., reverse the prediction) in the prediction/output. Counterfactual explanations can explain individual predictions. For instance, it can provide an explanation that describes causal situations such as “If A had not occurred, B would not have occurred”. Although counterfactual explanations are human-friendly, it suffers from the ``Rashomon effect'', where each counterfactual explanation tells a different story to reach a prediction. In other words, there are multiple true explanations (counterfactual) for each instance level prediction, and the challenge is how to choose the best one. The counterfactual methods do not require access to data or models and could work with a system that does not use machine learning at all. In addition, this method does not work well for categorical variables with many values. For instance, if the credit score of customer 5 (from Table \ref{tab:example-based}) can be increased to 749 (similar to the credit score of customer 6) from 748, given other features values remain unchanged, the customer will not default on a payment. In short, there can be multiple different ways to tune feature values to make customers move from non-default to default, or vice versa.

Traditional explanation methods are mostly based on explaining correlation rather than causation.  Moraffah et al. \cite{moraffah2020causal} focus on the causal interpretable model that explains the possible decision under different situations such as being trained with different inputs or hyperparameters. This causal interpretable approach share concept of counterfactual analysis as both work on causal inference. Their work also suggests possible use in fairness criteria evaluation of decisions. 
\begin{table}
\centering
\caption{Example-Based Explanations}
\label{tab:example-based}
\begin{tabular}{llll}
\toprule
Customer & Delinquency & Credit score & Defaulted  \\
\midrule
1 & 162                      & 680         & yes          \\
2 & 149                      & 691         & yes        \\
3 & 6                        & 728         & yes          \\
4 & 6                        & 744         & yes          \\
5 & 0                        & 748         & yes          \\
6 & 0                        & 749         & no          \\
7 & 0                        & 763         & no         \\
8 & 0                        & 790         & no         \\
9 & 0                        & 794         & no          \\
10 & 0                       & 806         & no         \\
\bottomrule
\end{tabular}
\end{table}

\subsubsection{Adversarial}
An adversarial technique is capable of flipping the decision using counterfactual examples to fool the machine learner (i.e., small intentional perturbations in input to make a false prediction). However, adversarial examples could help to discover hidden vulnerabilities as well as to improve the model. For instance, an attacker can intentionally design adversarial examples to cause the AI system to make a mistake (i.e., fooling the machine), which poses greater threats to cyber-security and autonomous vehicles. As an example, the credit default prediction system can be fooled for customer 5, just by increasing the credit score by 1 (see Table \ref{tab:example-based}), leading to a reversed prediction. 

Hartl et al. \cite{hartl2019explainability} emphasize on understanding the implications of adversarial samples on Recurrent Neural Network (RNNs) based IDS because RNNs are good for sequential data analysis, and network traffic exhibits some sequential patterns.  They find that adversarial the adversarial training procedure can significantly reduce the attack surface. Furthermore, \cite{marino2018adversarial} apply an adversarial approach to finding minimum modification of the input features of an intrusion detection system needed to reverse the classification of the misclassified instance. Besides satisfactory explanations of the reason for misclassification, their approach work provide further diagnosis capabilities.

\subsubsection{Prototypes}
Prototypes consist of a selected set of instances that represent the data very well. Conversely, the set of instances that do not represent data well are called criticisms\cite{kim2016examples}. Determining the optimal number of prototypes and criticisms are challenging. For example, customers 1 and 10 from Table \ref{tab:example-based} can be treated as prototypes as those are strong representatives of the corresponding target. On the other hand,  customers 5 and 6 (from Table \ref{tab:example-based}) can be treated as a criticism as the distance between the data points is minimal, and they might be classified under either class from run to run of the same or different models. 

\subsubsection{Influential Instances}
Influential instances are data points from the training set that are influential for prediction and parameter determination of the model. While it helps to debug the model and understand the behavior of the model better, determining the right cutoff point to separate influential or non-influential instances is challenging. For example, based on the values of feature credit score and delinquency, customers 1, 2, 9, and 10 from Table \ref{tab:example-based} can be treated as influential instances as those are strong representatives of the corresponding target. On the other hand, customers 5 and 6 are not influential instances, as those would be in the margin of the classification decision boundary. 

\subsubsection{k-nearest Neighbors Model}
\begin{figure}[h!]
    \centering
    \includegraphics[width=\linewidth]{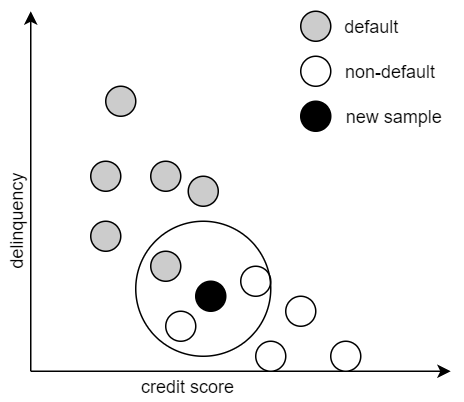}
    \caption{KNN}
    \label{fig:knn}
\end{figure}

The prediction of the k-nearest neighbor model can be explained with the k-neighbor data points (neighbors those were averaged to make the prediction). A visualization of the individual cluster containing similar instances provides an interpretation of why an instance is a member of a particular group or cluster. For example, in Figure \ref{fig:knn}, the new sample (black circle) is classified according to the other three (3-nearest neighbor) nearby samples(one gray, two white). This visualization gives an interpretation of why a particular sample is part of a particular class.

Table \ref{tab:interpretability_methods} summarizes the explainability methods from the perspective of (A) whether the method approximates the model behavior (i.e., creates an illusion of interpretability) or finds actual behavior, (B) whether the method alone is inherently interpretable or not, (C) whether the interpretation method is ante-hoc, that is, it incorporates explainability into a model from the beginning, or post-hoc, where explainability is incorporated after the regular training of the actual model (i.e., testing time), (D) whether the method is model agnostic (i.e., works for any ML model) or specific to an algorithm, and (E) whether the model is local, providing instance-level explanations, or global, providing overall model behavior. 

Our analysis says there is a lack of an explainability method (i.e., a gap in the literature), which is, at the same time actual and direct (i.e., does not create an illusion of explainability by approximating the model), model agnostic, and local, such that it utilizes the full potential of the explainability method in different applications. There are some recent works that bring external knowledge and infuse that into the model for better interpretation. These XAI methods have the potential to fill the gap to some extent by incorporating domain knowledge into the model in a model agnostic and transparent way (i.e., not by illusion).

\begin{table*}
\small
\centering
\caption{Comparison of different explainability methods from a set of key perspectives (approximation or actual; inherent or not; post-hoc or ante-hoc; model-agnostic or model specific; and global or local)}
\label{tab:interpretability_methods}
\begin{tabular}{p{6cm}lllll}
\toprule
Method                                   & Approx. & Inherent & Post/Ante & Agnos./Spec. & Global/Local  \\
\midrule
Linear/Logistic Regression               & No      & Yes      & Ante                 & Specific                   & Both             \\
Decision Trees                           & No      & Yes      & Ante                 & Specific                   & Both             \\
Decision Rules                           & No      & Yes      & Ante                 & Specific                   & Both             \\
k-Nearest Neighbors                      & No      & Yes      & Ante                 & Specific                   & Both             \\
Partial Dependence Plot (PDP)            & Yes     & No       & Post                 & Agnostic                   & Global           \\
Individual Conditional Expectation (ICE) & Yes     & No       & Post                 & Agnostic                   & Both             \\
Accumulated Local Effects (ALE) Plot     & Yes     & No       & Post                 & Agnostic                   & Global           \\
Feature Interaction                      & No      & Yes      & Both                 & Agnostic                   & Global           \\
Feature Importance                       & No      & Yes      & Both                 & Agnostic                   & Global           \\
Global Surrogate                         & Yes     & No       & Post                 & Agnostic                   & Global           \\
Local Surrogate (LIME)                   & Yes     & No       & Post                 & Agnostic                   & Local            \\
Shapley Values (SHAP)                    & Yes     & No       & Post                 & Agnostic                   & Local            \\
Break Down                    & Yes     & No       & Post                 & Agnostic                   & Local            \\
Counterfactual explanations              & Yes     & No       & Post                 & Agnostic                   & Local            \\
Adversarial examples                     & Yes     & No       & Post                 & Agnostic                   & Local            \\
Prototypes                               & Yes     & No       & Post                 & Agnostic                   & Local            \\
Influential instances                    & Yes     & No       & Post                 & Agnostic                   & Local            \\
\bottomrule
\end{tabular}
\end{table*}

\subsection{Other Techniques}
Chen et al. \cite{chen2018learning} introduce an instance-wise feature selection as a methodology for model interpretation where the model learns a function to extract a subset of most informative features for a particular instance. The feature selector attempt to maximize the mutual information between selected features and response variables. However, their approach is mostly limited to posthoc approaches. 

In a more recent work, \cite{jung2020information} study explainable ML using information theory where they quantify the effect of an explanation by the conditional mutual information between the explanation and prediction considering user background. Their approach provides personalized explanation based on the background of the recipient, for instance, a different explanation for those who know linear algebra and those who don’t. However, this work is yet to be considered as a comprehensive approach which considers a variety of user and their explanation needs.  
To understand the flow of information in a Deep Neural Network (DNN), \cite{ancona2017towards} analyzed different gradient-based attribution methods that assign an attribution value (i.e., contribution or relevance) to each input feature (i.e., neuron) of a network for each output neurons. They use a heatmap for better visualizations where a particular color represents features that contribute positively to the activation of target output, and another color for features that suppress the effect on it. 

A survey on the visual representation of Convolutional Neural Networks (CNNs), by \cite{zhang2018visual}, categorizes works based on a) visualization of CNN representations in intermediate network layers, b) diagnosis of CNN representation for feature space of different feature categories or potential representation flaws, c) disentanglement of “the mixture of patterns” encoded in each filter of CNNs, d) interpretable CNNs, and e) semantic disentanglement of CNN representations.

In the industrial control system, an alarm from the intrusion/anomaly detection system has a very limited role unless the alarm can be explained with more information. \cite{wang2020explaining} design a layer-wise relevance propagation method for DNN to map the abnormalities between the calculation process and features.  This process helps to compare the normal samples with abnormal samples for better understanding with detailed information. 

\subsection{Knowledge Infusion Techniques}
\cite{kim2017interpretability} propose a concept attribution-based approach (i.e., sensitivity to the concept) that provides an interpretation of the neural network's internal state in terms of human-friendly concepts. Their approach, \textit{Testing with CAV (TCAV)}, quantifies the prediction's sensitivity to a high dimensional concept. For example, a user-defined set of examples that defines the concept 'striped', TCAV can quantify the influence of 'striped' in the prediction of 'zebra' as a single number. However, their work is only for image classification and falls under the post-modeling notion (i.e., post-hoc) of explanation. 

\cite{kursuncu2019knowledge} propose a knowledge-infused learning that measures information loss in latent features learned by the neural networks through Knowledge Graphs (KGs). This external knowledge incorporation (via KGs) aids in supervising the learning of features for the model. Although much work remains, they believe that (KGs) will play a crucial role in developing explainable AI systems. 

\cite{islam2019infusing} and \cite{islam2019domain} infuse popular domain principles from the domain in the model and represent the output in terms of the domain principle for explainable decisions. In \cite{islam2019infusing}, for a bankruptcy prediction problem they use the 5C's of credit as the domain principle which is commonly used to analyze key factors: character (reputation of the borrower/firm), capital (leverage), capacity (volatility of the borrower's earnings), collateral (pledged asset) and cycle (macroeconomic conditions) \cite{angelini2008neural}, \cite{5cs_of_credit}. In \cite{islam2019domain}, for an intrusion detection and response problem, they incorporate the CIA principles into the model; \textit{C} stands for \textit{confidentiality}\textemdash{concealment of information or resources}, \textit{I} stands for \textit{integrity}\textemdash{trustworthiness of data or resources}, and \textit{A} stands for \textit{availability}\textemdash{ability to use the information or resource desired}  \cite{matt2006introduction}. In both cases, the infusion of domain knowledge leads to better explainability of the prediction with negligible compromises in performance. It also comes with better execution time and a more generalized model that works better with unknown samples. 

Although these works \cite{islam2019infusing, islam2019domain} come with unique combinations of merits such as model agnosticism, the capability of both local and global explanation, and authenticity of explanation\textemdash{simulation or emulation free}, they are still not fully off-the-shelf systems due to some domain-specific configuration requirements.  Much work still remains and needs further attention.

%
%
%
%################################### Challenges and Future Research Directions ############################ 
%
%
%
\section{Quantifying Explainability and Future Research Directions} \label{sec:challenge_future_research}
\subsection{Quantifying Explainability}
The quantification or evaluation of explainability is an open challenge. There are two primary directions of research towards the evaluation of explainability of an AI/ML model: (1) model complexity-based, and (2) human study-based.

\subsubsection{Model Complexity-based Explainability Evaluation}
In the literature, model complexity and (lack of) model interpretability are often treated as the same \cite{molnar2019quantifying}. For instance, in \cite{furnkranz2012rule},  \cite{yang2017scalable}, model size is often used as a measure of interpretability (e.g., number of decision rules, depth of the tree, number of non-zero coefficients). 

\cite{yang2017scalable} propose a scalable Bayesian Rule List (i.e., probabilistic rule list) consisting of a sequence of IF-THEN rules, identical to a decision list or one-sided decision tree. Unlike the decision tree that uses greedy splitting and pruning, their approach produces a highly sparse and accurate rule list with a balance between interpretability, accuracy, and computation speed. Similarly, the work of  \cite{furnkranz2012rule} is also rule-based. They attempt to evaluate the quality of the rules using a rule learning algorithm by: the observed coverage, which is the number of positive examples covered by the rule, which should be maximized to explain the training data well; and consistency, which is the number of negative examples covered by the rule, which should be minimized to generalize well to unseen data. 

According to \cite{ruping2006learning}, while the number of features and the size of the decision tree are directly related to interpretability, the optimization of the tree size or features (i.e., feature selection) is costly as it requires the generation of a large set of models and their elimination in subsequent steps. However, reducing the tree size (i.e., reducing complexity) increases error, as they could not find a way to formulate the relation in a simple functional form. 
More recently, \cite{molnar2019quantifying} attempts to quantify the complexity of the arbitrary machine learning model with a model agnostic measure. In that work, the author demonstrates that when the feature interaction (i.e., the correlation among features) increases, the quality of representations of explainability tools degrades. For instance, the explainability tool ALE Plot (see Figure \ref{fig:ale} starts to show harsh lines (i.e., zigzag lines) as feature interaction increases. In other words, with more interaction comes a more combined influence in the prediction, induced from different correlated subsets of features (at least two), which ultimately makes it hard to understand the causal relationship between input and output, compared to an individual feature influence in the prediction. In fact, from our study of different explainability tools (e.g., LIME, SHAP, PDP), we have found that the correlation among features is a key stumbling block to represent feature contribution in a model agnostic way. Keeping the issue of feature interactions in mind, \cite{molnar2019quantifying} propose a technique that uses three measures: number of features, interaction strength among features, and the main effect (excluding the interaction part) of features, to measure the complexity of a post-hoc model for explanation. 

Although,\cite{molnar2019quantifying} mainly focuses on model complexity for post-hoc models, their work was a foundation for the approach by \cite{islam2019towards} for the quantification of explainability. Their approach to quantify explainability is model agnostic and is for a model of any notion (e.g., pre-modeling, post-hoc) using proxy tasks that do not involve a human. Instead, they use known truth as a metric (e.g., the less number of features, the more explainable the model). Their proposed formula for explainability gives a score in between 0 and 1 for explainability based on the number of cognitive chunks (i.e., individual pieces of information) used on the input side and output side, and the extent of interaction among those cognitive chunks. 

\subsubsection{Human Study-based Explainability Evaluation}
The following works deal with the application-level and human-level evaluation of explainability involving human studies.

\cite{huysmans2011empirical} investigate the suitability of different alternative representation formats (e.g., decision tables, (binary) decision trees, propositional rules, and oblique rules) for classification tasks primarily focusing on the explainability of results rather than accuracy or precision. They discover that decision tables are the best in terms of accuracy, response time, the confidence of answer, and ease of use. 

\cite{dhurandhar2017tip} argue that interpretability is not an absolute concept; instead, it is relative to the target model, and may or may not be relative to the human. Their finding suggests that a model is readily interpretable to a human when it uses no more than seven pieces of information \cite{miller1956magical}. Although, this might vary from task to task and person to person. For instance, a domain expert might consume a lot more detailed information depending on their experience. 

The work of \cite{poursabzi2018manipulating} is a human-centered approach, focusing on previous work on human trust in a model from psychology, social science, machine learning, and human-computer interaction communities. In their experiment with human subjects, they vary factors (e.g., number of features, whether the model internals are transparent or a black box) that make a model more or less interpretable and measures how the variation impacts the prediction of human subjects. Their results suggest that participants who were shown a transparent model with a small number of features were more successful in simulating the model's predictions and trusted the model's predictions. 

\cite{friedler2019assessing} investigate interpretability of a model based on two of its definitions: simulatability, which is a user's ability to predict the output of a model on a given input; and ``what if'' local explainability, which is a user's ability to predict changes in prediction in response to changes in input, given the user has the knowledge of a model's original prediction for the original input. They introduce a simple metric called \textit{runtime operation count} that measures the interpretability, that is, the number of operations (e.g., the arithmetic operation for regression, the boolean operation for trees) needed in a user's mind to interpret something. Their findings suggest that interpretability decreases with an increase in the number of operations. 

Despite some progress, there are still some open challenges surrounding explainability such as an agreement of what an explanation is and to whom; a formalism for the explanation; and quantifying the human comprehensibility of the explanation. Other challenges include addressing more comprehensive human studies requirements and investigating the effectiveness among different approaches (e.g., supervised, unsupervised, semi-supervised) for various application areas (e.g., natural language processing, image recognition). 

\subsection{Future Research Directions}
The long term goal for current AI initiatives is to contribute to the design, development, and deployment of human-centered artificial intelligent systems, where the agents collaborate with the human in an interpretable and explainable manner, with the intent on ensuring fairness, transparency, and accountability. To accomplish that goal, we propose a set of research plans/directions towards achieving responsible or human-centered AI using XAI as a medium.

\subsubsection{A Generic Framework to Formalize Explainable Artificial Intelligence}
The work in \cite{islam2019infusing} and \cite{islam2019domain}, demonstrates a way to collect and leverage domain knowledge from two different domains, finance and cybersecurity, and further infused that knowledge into black-box models for better explainability. In both of these works, competitive performance with enhanced explainability is achieved. However, there are some open challenges such as (A) a lack of formalism of the explanation, (B) a customized explanation for different types of explanation recipients (e.g., layperson, domain expert, another machine), (C) a way to quantify the explanation, and (D) quantifying the level of comprehensibility with human studies. Therefore, leveraging the knowledge from multiple domains, a generic framework could be useful considering the mentioned challenges. As a result, mission-critical applications from different domains will be able to leverage the black-box model with greater confidence and regulatory compliance. 

\subsubsection{Towards Fair, Accountable, and Transparent AI-based Models}
Responsible use of AI is crucial for avoiding risks stemming from a lack of fairness, accountability, and transparency in the model. Remediation of data, algorithmic, and societal biases is vital to promote fairness; the AI system/adopter should be held accountable to affected parties for its decision; and finally, an AI system should be analyzable, where the degree of transparency should be comprehensible to have trust in the model and its prediction for mission-critical applications. Interestingly, XAI enhances understating directly, increasing trust as a side-effect. In addition, the explanation techniques can help in uncovering potential risks (e.g., what are possible fairness risks). So it is crucial to adhere to fairness, accountability, and transparency principles in the design and development of explainable models. 

\subsubsection{Human-Machine Teaming}
To ensure the responsible use of AI, the design, development, and deployment of human-centered AI, that collaborates with the humans in an explainable manner, is essential. Therefore, the explanation from the model needs to be comprehensible by the user, and there might be some supplementary questions that need to be answered for a clear explanation. So, the interaction (e.g., follow-ups after the initial explanation) between humans and machines is important. The interaction is more crucial for adaptive explainable models that provide context-aware explanations based on user profiles such as expertise, domain knowledge, interests, and cultural backgrounds. The social sciences and human behavioral studies have the potential to impact XAI and human-centered AI research. Unfortunately, the Human-Computer Interaction (HCI) community is kind of isolated. The combination of HCI empirical studies and human science theories could be a compelling force for the design of human-centered AI models as well as furthering XAI research. Therefore, efforts to bring a human into the loop, enabling the model to receive input (repeated feedback) from the provided visualization/explanations to the human, and improving itself with the repeated interactions, has the potential to further human-centered AI. Besides adherence to fairness, accountability, and transparency, the effort will also help in developing models that adhere to our ethics, judgment, and social norms.  

\subsubsection{Collective Intelligence from Multiple Disciplines}
From the explanation perspective, there is plenty of research in philosophy, psychology, and cognitive science on how people generate, select, evaluate, and represent explanations and associate cognitive biases and social expectations in the explanation process. In addition, from the interaction perspective, human-computer teaming involving social science, the HCI community, and social-behavioral studies could combine for further breakthroughs. Furthermore, from the application perspective, the collectively learned knowledge from different domains (e.g., Health-care, Finance, Medicine, Security, Defense) can contribute to furthering human-centric AI and XAI research. Thus, there is a need for a growing interest in multidisciplinary research to promote human-centric AI as well as XAI in mission-critical applications from different domains.    

%
%
%
%###################################  CONCLUSION ############################ 
%
%
%
\section{Conclusion}\label{sec:conclusion}
We demonstrate and analyze mutual XAI methods using a mutual test case to explain competitive advantages and elucidate the challenges and further research directions. Most of the available works on XAI are on the post-hoc notion of explainability. However, the post-hoc notion of explainability is not purely transparent and can be misleading, as it explains the decision after it has been made. The explanation algorithm can be optimized to placate subjective demand, primarily stemming from the emulation effort of the actual prediction, and the explanation can be misleading, even when it seems plausible \cite{lipton2016mythos,on_explainable_artificial_intelligence}. Thus, many suggest not to explain black-box models using post-hoc notions, instead, they suggest adhering to simple and intrinsically explainable models for high stakes decisions \cite{rudin2019stop}. Furthermore, from the literature review, we find that explainability in pre-modeling is a viable option to avoid the transparency related issues, albeit, under-focused. In addition, knowledge infusion techniques have the potential to enhance explainability greatly, although, also an under-focused challenge. Therefore, we need more focus on the explainability of ``black box'' models using domain knowledge. At the same time, we need to focus on the evaluation or quantification of explainability using both human and non-human studies. We believe this review provides a good insight into the current progress on XAI approaches, evaluation and quantification of explainability, open challenges, and a path towards responsible or human-centered AI using XAI as a medium.

% use section* for acknowledgment
\ifCLASSOPTIONcompsoc
  % The Computer Society usually uses the plural form
  \section*{Acknowledgments}
\else
  % regular IEEE prefers the singular form
  \section*{Acknowledgment}
\fi
Our sincere thanks to Christoph Molnar for his open E-book on Interpretable Machine Learning and contribution to the open-source R package ``iml''. Both were very useful in conducting this survey.

\bibliographystyle{IEEEtran}
\bibliography{bib}

\end{document}